\documentclass{article}

\usepackage{other/arxiv}

\usepackage[utf8]{inputenc} 
\usepackage[T1]{fontenc}    
\usepackage{hyperref}       
\usepackage{url}            
\usepackage{booktabs}       
\usepackage{amsfonts}       
\usepackage{nicefrac}       
\usepackage{microtype}      
\usepackage{lipsum}
\usepackage{graphicx}
\usepackage{subcaption}
\usepackage{mathtools, nccmath}
\usepackage{enumerate}

\newcommand{\dar}{\textit{AR-Net}}
\newcommand{\car}{\textit{Classic-AR}}
\newcommand{\NN}{\textit{FF-NN}}

\title{\dar: A simple Auto-Regressive Neural Network \\ for time-series} 

\author{
    Oskar J.~Triebe\\
    Stanford University\\
    \texttt{triebe@stanford.edu}
    \And
    Nikolay Laptev\\
    Facebook, Inc.\\
    \texttt{nlaptev@fb.com}
    \And
    Ram Rajagopal\\
    Stanford University\\
    \texttt{ramr@stanford.edu}
} 

\begin{document}
\maketitle
\begin{abstract}
    In this paper we present a new framework for time-series modeling that combines the best of traditional statistical models and neural networks.
    We focus on time-series with long-range dependencies, needed for monitoring fine granularity data (e.g. minutes, seconds, milliseconds), prevalent in operational use-cases.
    
    Traditional models, such as auto-regression fitted with least squares (\car) can model time-series with a concise and interpretable model.  When dealing with long-range dependencies, \car~ models can become intractably slow to fit for large data. Recently, sequence-to-sequence models, such as Recurrent Neural Networks, which were originally intended for natural language processing, have become popular for time-series. However, they can be overly complex for typical time-series data and lack interpretability.
    
    A scalable and interpretable model is needed to bridge the statistical and deep learning-based approaches.
    As a first step towards this goal, we propose modelling AR-process dynamics using a feed-forward neural network approach, termed \dar. We show that \dar~ is as interpretable as \car~ but also scales to long-range dependencies.
    
    Our results lead to three major conclusions: 
    First, \dar~ learns identical AR-coefficients as \car, thus being equally interpretable. 
    Second, the computational complexity with respect to the order of the AR process, is linear for \dar~ as compared to a quadratic for \car. 
    This makes it possible to model long-range dependencies within fine granularity data.
    Third, by introducing regularization, \dar~ automatically selects and learns sparse AR-coefficients. 
    This eliminates the need to know the exact order of the AR-process and allows to learn sparse weights for a model with long-range dependencies.
\end{abstract}
\section{Introduction}
\label{introducetion}

Temporal data has an intrinsic time component that is present in most real-world applications (i.e., sensor measurements, the stock-market etc). 
A time series is a sequence of time-ordered data values that measures some process. 
Time-dependent data-center traffic is an example of a time-series. 

Forecasting is an important task in time-series application. 
Forecasting can be for several years in advance for organizational planning or a few seconds ahead for operational automation. 
To have an accurate forecast, we have to understand whether there are any factors that influence the process, and whether the process can influence itself.
We set the scope of this paper to fully auto-regressive time-series and focus on medium to large volume data.

\paragraph{Auto-Regression}
While many advanced forecasting methods have previously been developed, we focus on the more fundamental and most commonly used auto-regression (AR) based time-series models.  
Auto-regressive models are remarkably flexible at handling a wide range of different time series patterns \cite{hyndman2014forecasting} and have been widely used in practice.
Statistical models exploit the inherent characteristics of a time series, leading to a concise model. 
This is possible because the model makes strong assumptions about the data, such as the true order of the AR-process.
The order $p$ of an AR($p$) process is defined as the number of previous values of the time-series (lags) upon which the next value is dependent.
\textit{AR} processes with a high $p$-order are important for monitoring fine granularity data (e.g., minutes, seconds, milliseconds), and for long-range dependencies, where values long past still influence future outcomes.
Prevalent operational use-cases are data centers, wireless networks or Internet of Things (IoT) applications \cite{Ma_2017, wireless_neural}.
The parameters of an AR model are traditionally fitted using least squares (\car). 
Unfortunately, when modelling long-range dependencies the fitting procedure of \car~ models with a large $p$-order can become impractically slow, as we will demonstrate in this paper. 

\paragraph{Neural Networks}
To overcome the scalability challenges, the time-series community has started to adopt deep learning methods such as Recurrent Neural Networks (RNN) and Convolutional Neural Networks (CNN). However, in their current form, RNNs and CNNs are designed for rich natural language processing or imaging data, making them too complex for most time-series applications. Their adoption has been further limited by the difficulty to make the models explainable to decision stakeholders.

However, there are two attributes which make general neural networks attractive for time-series modeling. 
First, neural networks have general nonlinear function mapping capability which can approximate any continuous function. 
Hence, it is capable of solving many complex problems, given adequate data. 
Second, a neural network is a non-parametric data-driven model and it does not require restrictive assumptions on the underlying process from which data are generated. Because of this feature, it is less susceptible to model mis-specification problems compared to  most parametric nonlinear methods \cite{Hornik:1989:MFN:70405.70408, Cybenko1989}. 
This is an important advantage since time-series modeling does not show a specific nonlinear pattern. Different time-series may have unique behavior not captured by a parametric model.

\paragraph{In this paper} We discuss the parallels between Feed-Forward Neural Networks (\NN) and \car~ models to overcome scalability issues, while ensuring interpretability and model simplicity.
We use auto-regression to explain the system dynamics and learn the parameters with a neural network, termed  \dar.
While designing \dar~ we focused on model simplicity and interpretability. 
Our goal was to re-introduce the time-series community to deep learning by providing an interpretable, fast and easy to use alternative to \car. Specifically, in this paper we show that:
\begin{itemize}
    \item \dar~ in its basic form is as interpretable as \car, as they learn near identical parameters.
    \item \dar~ scales well to large $p$-orders, making it possible to estimate long-range dependencies (important in high resolution monitoring applications).
    \item \dar~ automatically selects and estimates the important coefficients of a sparse AR process, thus eliminating the need to know the true order of the AR process.
\end{itemize}

\paragraph{Paper structure} 
In section~\ref{related}, we review relevant previous work. 
Next, we describe how we introduced AR dynamics and sparsity into the model in section~\ref{methods}.
In section~\ref{results}, we compare \dar~ and \car~ model performance and behavior on time series data. 
Finally, we summarize our work and vision for future work in section~\ref{conclusion}. 
We include a review of neural networks and fitting procedures as an appendix in section~\ref{appendix}.
\section{Related Work}
\label{related}
Models such as (S)ARIMA(X) and Prophet\cite{prophetarticle}, exploit the inherent characteristics of a time series, leading to a more concise model. This is possible because the model makes strong assumptions about the data, such as the true order of the AR-process, the trend or the seasonality. These models, however, do not scale well for a large volume of training data and are hard to extend, particularly if there are long-range dependencies or complex interactions~\cite{Sutskever:2014:SSL:2969033.2969173}.

To overcome the scalability challenges, `sequence-to-sequence' deep learning methods based on recurrence or convolution, have been successfully developed in natural language processing. The most prominent examples being recurrent neural networks (RNN), such as the long short term memory cell (LSTM)\cite{LSTM}, attention~\cite{attention} or convolution based approaches such as Wavenet~\cite{wavenet}.
Recently, they have also become popular in the time-series community, as they allow for a more expressive model without the need to engineer elaborate features~(\cite{NIPS2014_5346}, \cite{cond_wavenet},  \cite{DBLP:journals/corr/CuiCC16}, \cite{DBLP:journals/corr/abs-1903-02540}). 

While these models scale well to applications with rich data, they can be overly complex for typical time-series data. 
Another reason why RNNs have not been widely adopted for time-series applications is that they are generally regarded as ``black boxes'' by practitioners.
Though their complex parameter interactions are interpretable to some extent (e.g., LIME\cite{LIME}, SHAP\cite{SHAP}), they are difficult to interpret compared to AR-based models, restricting their adoption in practice where model explainability is key.

It has become increasingly accessible for practitioners to train deep learning models with a framework like Keras~\cite{chollet2015keras} without actually understanding the model dynamics. As a consequence, some research papers blindly apply a popular model to a time-series problem, without evaluating whether it is suited for the application. The resulting conclusions have little to no meaning, often mis-leading other practitioners to a wrong understanding of deep-learning. An example is \cite{badexample}, where the authors compare a host of models, among which an LSTM, evaluated on time-series with ten to one hundred entries. With standard Keras parameters, the LSTM is over-parametrized by multiple magnitudes, leading to a meaningless comparison with simpler statistical models. The authors conclude that statistical models are better for time-series modelling. Other similar work \cite{10.1371/journal.pone.0194889,10.1371/journal.pone.0211057,DBLP:journals/corr/ChePCSL16} point to the need of adequate and easy-to-understand deep learning method for time-series modelling.

The closest work to ours is that of \cite{Tang93feed-forwardneural}, where the authors have compared neural networks as models for time series forecasting with those of the Box-Jenkins methods for long and short term memory series. 
The experiments indicate that for time series with long memory, both methods produced comparable results. 
However, for series with short memory, neural networks outperformed the Box-Jenkins model. Because neural networks can be easily built for multiple-step-ahead forecasting, they may present a better long term forecast model than the Box-Jenkins method. 
We add to their work by explicitly drawing the parallels between $AR(p)$ while focusing on concrete benefits of a neural network, targeting scalability and ease of use.

\paragraph{Our Contribution}
The majority of the time-series related literature focuses on complex models. 
Our research, however, makes a more fundamental observation about the differences between 
neural networks and the simplest classical auto-regressive (AR) model. 
We focus on simple feed forward neural networks to promote explainability and 
simplicity that parallels classical time-series models with the added benefits of scalability. 
We intentionally did not use more powerful methods, such as modeling latent states with recurrent networks or convolution because our goal was to bridge, not widen, the gap between traditional time-series and deep learning methods. We hope to show with \dar~ that deep learning models can be simple, interpretable, fast and easy to use, so that the time-series community may consider deep learning a viable option.

We formulate a simple neural network that mimics the \car~model, with the only difference being how they are fitted to data. Our model termed \dar, in it's simplest form, is identical to linear regression, fitted with stochastic gradient descent (SGD).
We show that \dar~ is identically interpretable as a \car~model and scales to large $p$-orders. 
As we discuss in the future work section, our vision is to leverage more powerful temporal modeling techniques of deep learning \textit{without} sacrificing  interpretability via explicit modeling of time-series components.
\section{Methods}
\label{methods}

\subsection{Data}
As we need knowledge of the true underlying AR coefficients in order to quantitatively evaluate the quality of the fitted weights, we use synthetic data. 
We generate the data with a noisy AR-process. 
For each run, random AR coefficients are sampled and a new time-series is generated with random normal noise of one standard deviation. The sampled weights are scaled to $(\sum_{i=1}^{i=p}{|w_i|}) \leq 1$.
Unless otherwise specified, the generated time-series is 125,000 samples long, split into 100,000 samples for training and the last 25,000 samples for testing.

In the sparse AR experiments, we fix the AR-parameters to $[0.2, 0.3, -0.5]$ in order to reduce randomness introduced by possibly unstable AR-parameter combinations. We still generate each time-series with new random noise. As most neural-network based methods are known to need large datasets, we further do some special experiments on a time-series with only 1000 samples for training and 1000 samples for testing to demonstrate that our method is also suited for medium sized datasets.

\subsection{\car~ Model}
\label{car}
In an auto-regressive model, we forecast the variable of interest using a linear combination of past values of the variable. The term ``auto'' in Auto-Regression (AR) indicates that the variable is regressed against itself \cite{hyndman2014forecasting}. This is like a multiple regression but with lagged values of the time-series $y_t$ as predictors. 
We refer to this as an AR$(p)$ model, an auto-regressive model of order $p$. An AR model of order  $p$ can be written as:
$$ y_t=c+\sum_{i=1}^{i=p}{w_i*y_{t-i}}+e_t $$
Where $y_{t-1, ..., t-p}$ are the $p$ lag terms used to predict $y_t$ and $\epsilon_t$ is white noise. 
The $p$ weights $w_i$, by which each of the $p$ lags $y_{t-i}$ is multiplied, are also referred to as the AR-coefficients.
As baseline, we use a traditional implementation of the auto-regressive model, fitted with least squares. We will refer to this model as \car.

\subsection{\dar~ Model}
\label{dar}
\begin{figure}[ht]
\centering
\begin{subfigure}{.5\textwidth}
  \centering
  \includegraphics[height=0.5\linewidth]{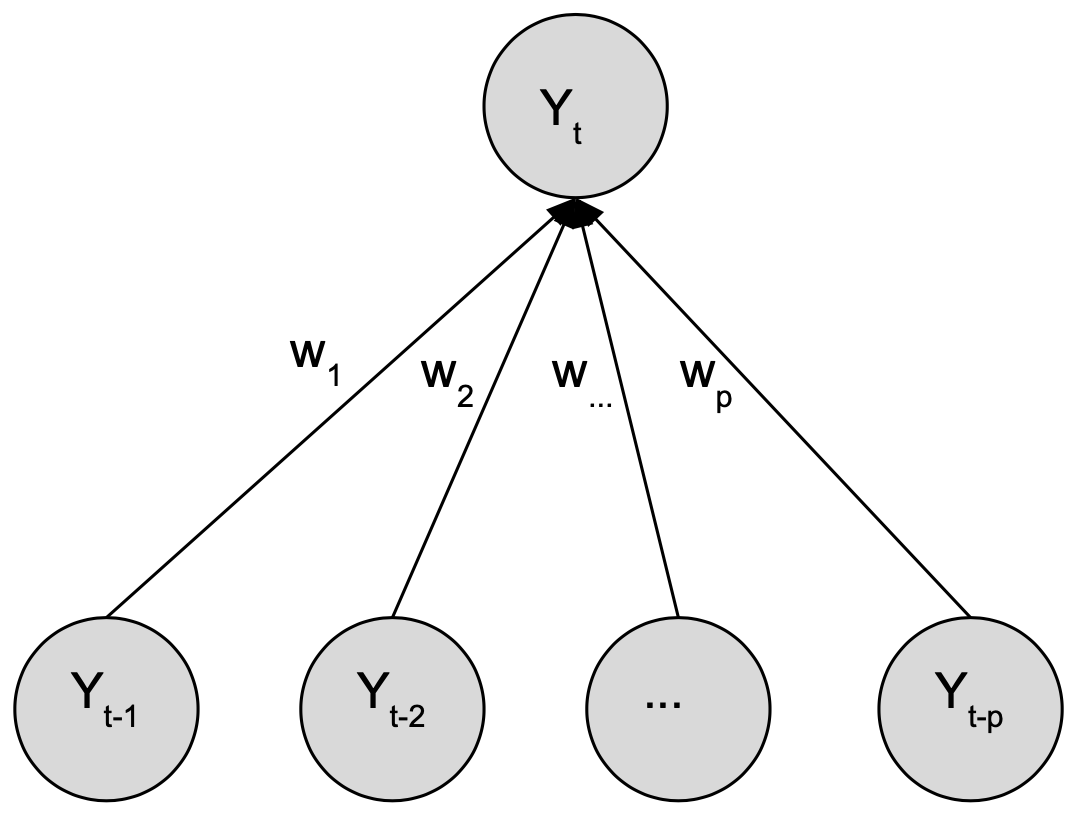} 
  \caption{An AR equivalent neural network architecture. The node weights $w_1, w_2, ... w_p$ correspond to the AR-coefficients.}
  \label{fig:argraph}
\end{subfigure}%
\begin{subfigure}{.5\textwidth}
  \centering
  \includegraphics[height=0.5\linewidth]{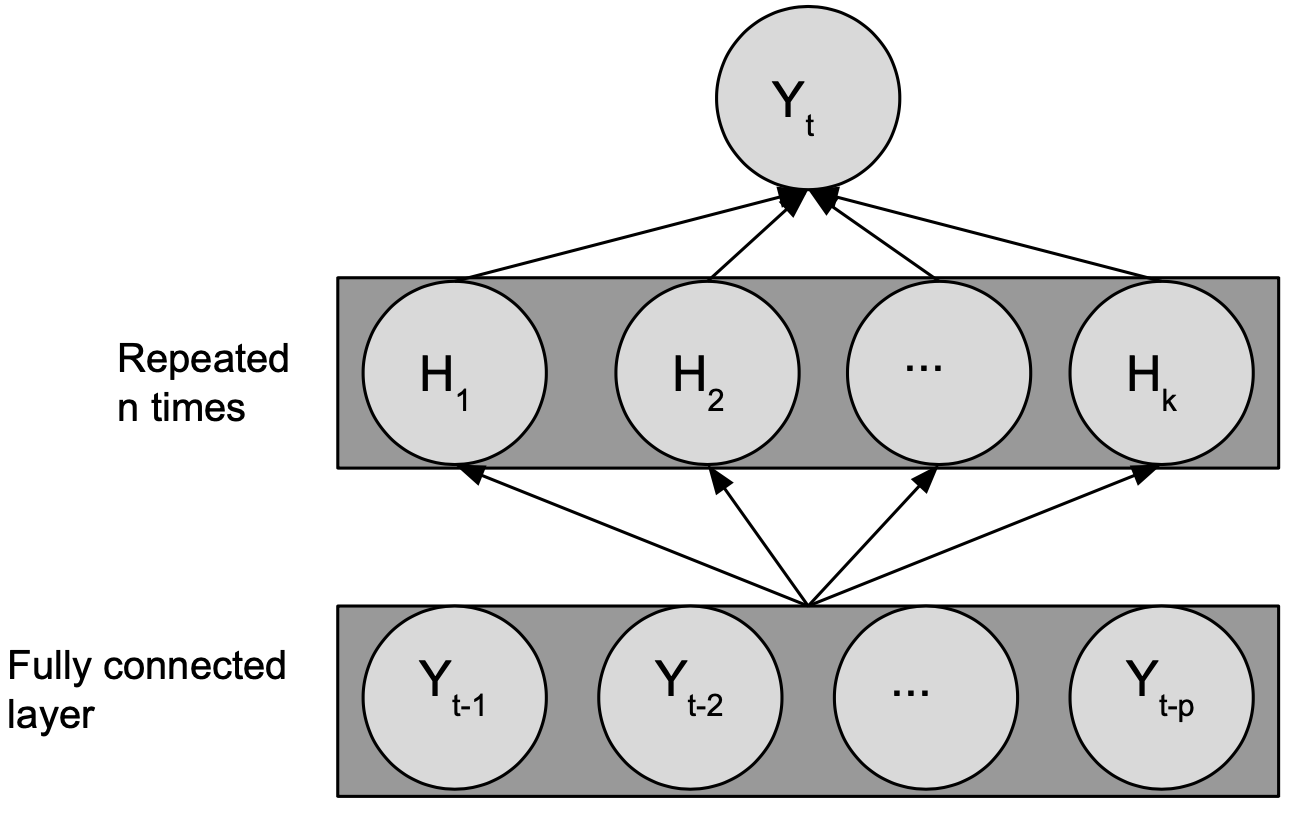}
  \caption{An AR inspired neural network \\ with n hidden layers of size $k$}
  \label{fig:dargraph}
\end{subfigure}
\caption{Schematic of \dar~ architectures with the lags $y_{t-1}, ..., y_{t-p}$ as inputs and $y_{t}$ as target.}
\label{fig:graphs}
\end{figure}

We propose \dar~ which mimics the traditional AR process with a neural network. It is designed such that that the parameters of its first layer are equivalent to the AR-coefficients (see figure \ref{fig:argraph}).
\dar~ can optionally be extended with hidden layers to achieve greater forecasting accuracy, at the cost of direct interpretability (see figure \ref{fig:dargraph}). In this paper, we will only evaluate the \dar~ model structure without hidden layers. 

In order to fit the \dar~ model to the same objective as \car~ optimized by least squares, we define the loss term to be Mean Squared Error (MSE). We could use any other loss here, such as the Hinge-loss, but for the sake of comparability with \car, we use MSE:
$$ min_{\theta}~~~L(y, \hat{y}, \theta) = \frac{1}{n} \sum_1^n{(y - \hat{y_{\theta}})^2} $$

\subsubsection{Sparse \dar}
In order to relax the constraint of knowing the true AR order, we can fit a larger model with sparse AR coefficients. This will also do away with the assumption that the AR-coefficients must consist of consecutive lags. We achieve this by adding a regularization term $R$ to the loss $L$ being minimized. 
$$ min_{\theta}~~~L(y, \hat{y}, \theta) + \lambda(s) \cdot R(\theta)$$
$$ \lambda(s) = c_{\lambda} \cdot (s^{-1} -1) $$

With parameters: \\\\
$ s = \frac{~ \hat{p}_{data}}{p_{model}} \text{~~the estimated sparsity of the AR coefficients, user-defined.} $ \\
$ c_{\lambda} \approx \frac{\sqrt{\hat{L}}}{100} \text{~~ Regularization strength, depends on estimated noise of data.}$ \\

For normalized data, the only parameter to be set is the estimated or desired sparsity $s$ of the AR-coefficients. The regularization strength $c_{\lambda}$ can be set manually or simply as a function of the estimated noise standard deviation $\sqrt{\hat{L}}$.
We experimented with different regularization functions $ R(\theta)$, including the known $L1$-regularization (``Lasso''). However, our regularization objective is different from most applications. We do not want to discourage large weights, like an $L1$-norm or $L2$-norm would. Instead we want to encourage the optimizer to set small weights to zero while keeping the other weights untouched. For us it is important that the actual weights are not regularized to be smaller than their unregularized optima, as they actually represent the AR coefficients. Our regularization function achieves this by having a large gradient close to zero and then quickly decreasing closer to one. Like this, the gradients of regularized weights further from zero basically vanish. We achieve this behavior by using a modified combination of a root and sigmoid transform of the absolute weight values:

$$ R(\theta) = \frac{1}{p} \sum_{i=1}^p{\frac{2}{1 + \exp(-c_{1} \cdot |\theta_i|^{\frac{1}{c_{2}}})} - 1} $$ 

The regularization curve parameters $c_{1}, c_{2} $, depend on the AR-coefficients range. For normalized data, with AR-coefficients in range $[0, 1]$, $ c_{1} \approx 3$ and $c_{2} \approx 3$ work ideally. For unnormalized data, we found a regularization function composed of a simple square root transform to work well. However, it does penalize larger weights unnecessarily. We include it here for completeness, though we did not use it in the presented results:
$$ R_{alt}(\theta) = \frac{1}{p} \sum_{i=1}^p{\sqrt{|\theta_i|}} $$

\subsection{Metrics}
We compared the models based on two metrics. Primarily, we evaluated the precision of the fitted AR-coefficients compared to the true AR-coefficients of the AR-process used to generate the data. Secondarily, we evaluated their one-step-ahead forecast performance.

The precision of the fitted AR-coefficients is measured by their symmetrical Total Percentage Error (sTPE), defined as:
$$ sTPE = 100 \cdot \frac{\sum_{i=1}^{i=p}{|\hat{w_i} - w_i|}}{\sum_{i=1}^{i=p}{|\hat{w_i}| + |w_i|}} $$
where $\hat{w}$ depicts the fitted AR-coefficients by the model and $w$ the true AR-coefficients.

The Mean Squared Error (MSE) of the one-step ahead forecast is given by:
$$ MSE = \frac{1}{n} \sum_1^n{(y_t - \hat{y_t})^2} $$
where $\hat{y_t}$ depicts the predicted next value by the model and $y_t$ the true next value of the time-series.
\section{Results}
\label{results}

In our experiments, we compared \dar~ and \car~ based on precision of their fitted AR-coefficients (sTPE). We made sure, that their on forecasting performance (MSE) remained close to ideal. Due to computational time constraints, experiments for \car~ with $p>25$ could not be conducted.

\subsection{Learning AR-Coefficients}
\label{sec:coeff_exp}
The first set of experiments examined whether \dar~ can accurately learn the dynamics of an AR-process. 
For each experiment, we produced ten noisy AR-process time series for each order of $p$. We then estimated the AR-parameters via classic least squares and via SGD on the same time-series. 
Note that here we assume to have knowledge of the true p-order of the AR-process. 

\begin{figure}[htbp]
\centering
\begin{subfigure}{0.495\textwidth}
  \centering
  \includegraphics[height=0.7\linewidth]{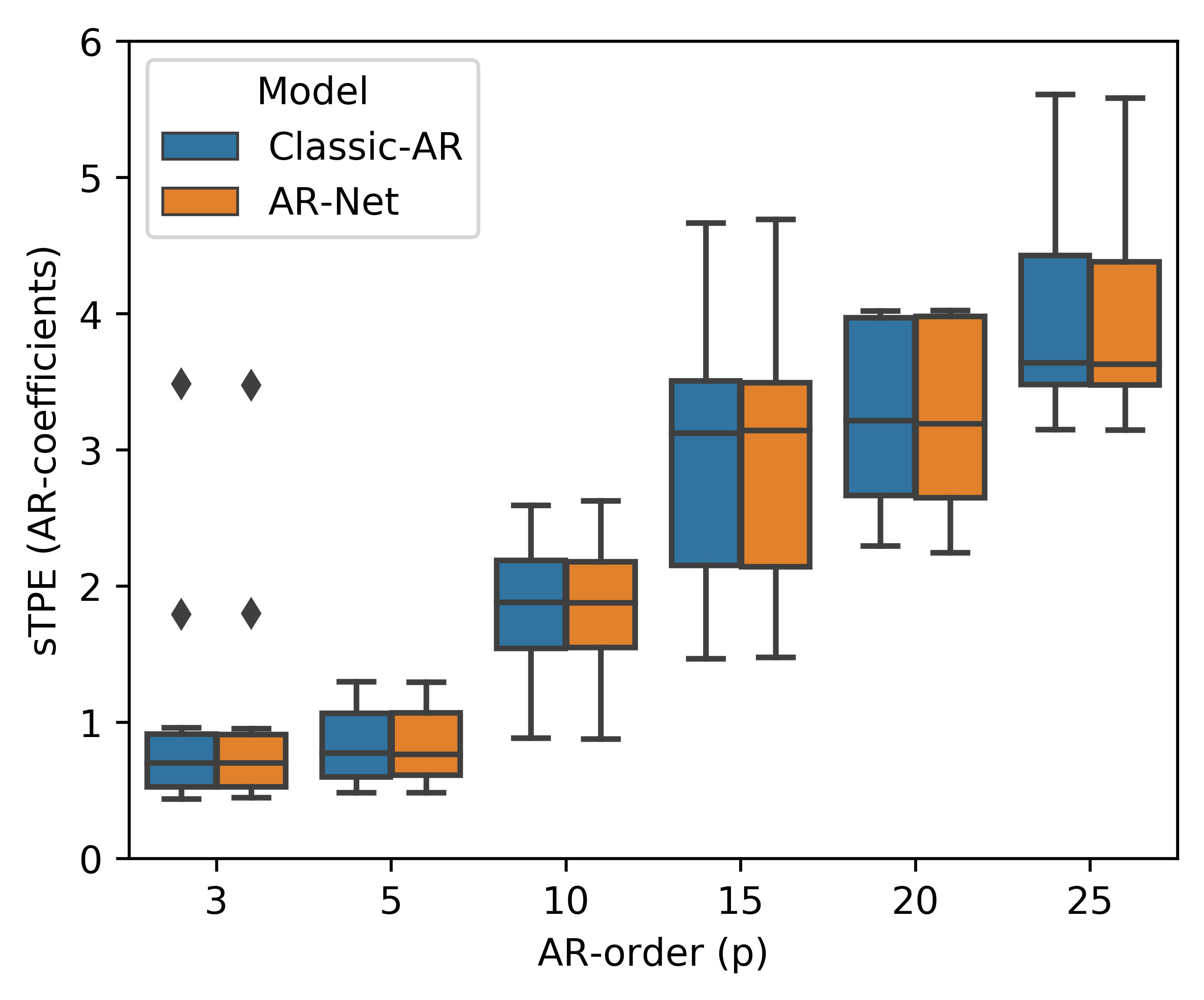}
  \caption{The precision of learned weights of \dar~ were identical to those of classic AR. The error of both methods monotonically  increased with $p$, with the exception of $p=1$.}
  \label{fig:full_sTPE}
\end{subfigure}
\begin{subfigure}{0.495\textwidth}
  \centering
  \includegraphics[height=0.7\linewidth]{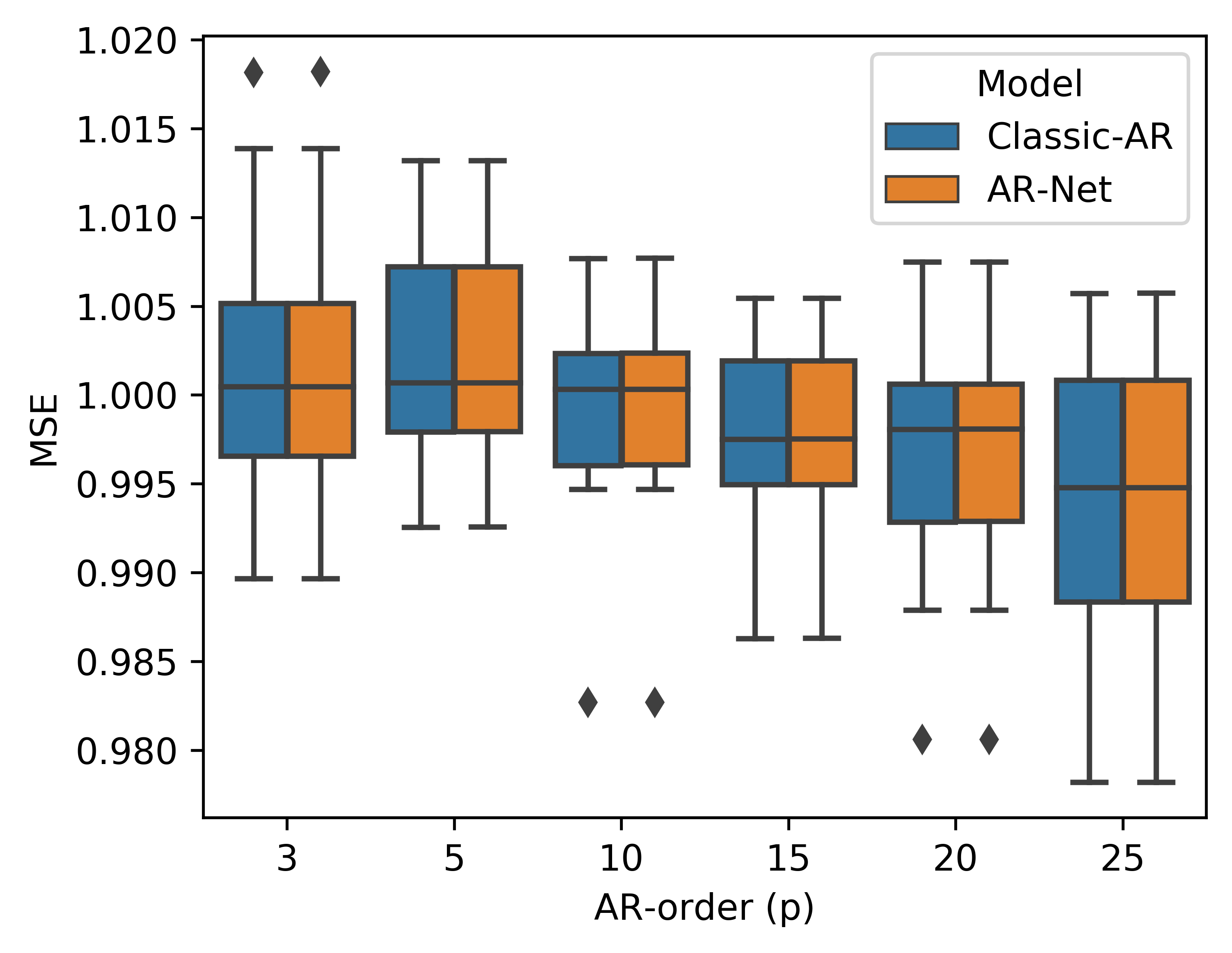}
  \caption{\dar~ and \car~ both effectively minimized the one-step-ahead forecasting MSE loss term.}
  \label{fig:full_MSE}
\end{subfigure}
\caption{\dar~ and \car~ both effectively learned the true AR-process coefficients, while minimizing the mean squared prediction error.}
\label{fig:results-1-25}
\end{figure}

Figure~\ref{fig:full_sTPE} shows that the precision of weights learned by \dar~ is identical to AR.
In terms of forecasting performance (MSE), we found no difference between the \car~ and \dar~ results. 
Though the fitting mechanisms were different, both algorithms were fitted to minimize the squared error of a one-step-ahead forecast.
In figure~\ref{fig:full_MSE} we can see that the mean squared errors of both models were near 1.0, same as the amount of noise in the time-series. 

\begin{figure}[htbp]
\centering
\includegraphics[width=.8\linewidth]{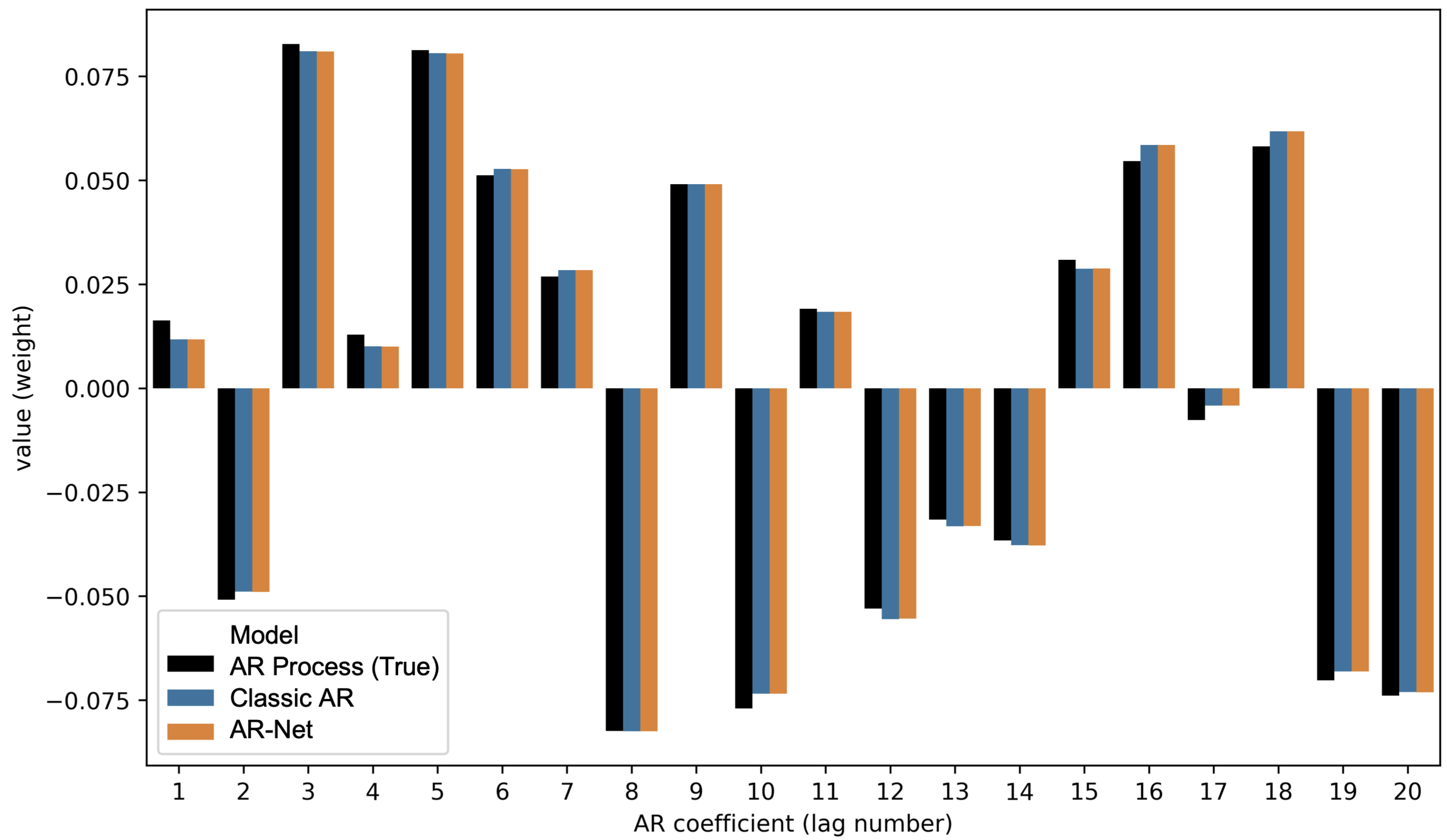}
\caption{Both \car~ and \dar~ learn AR coefficients which are qualitatively close to the true underlying system dynamics. Fitted on an AR(20)-process with noise of one standard deviation.}
\label{fig:weights}
\end{figure}

An example of the AR coefficients learned via SGD (\dar) and least squares estimation (\car) is shown in figure~\ref{fig:weights}. 
A qualitative analysis of the residuals by plotting the errors against their underlying value showed a similar near-Gaussian fit for both models (not shown).

\newpage
\subsubsection{Sparse AR coefficients for unknown p-order}
The regularized \dar~ model of $p$-order works for any smaller order AR-process without the need to know the real order precisely (up to one and a half magnitudes). 
In our experimental setups we used a non-sparse but smaller order AR-process to demonstrate the ability to fit sparse AR-coefficients. 
Both \car~ and \dar~ models were trained on data generated by an AR-3 process with noise of one standard deviation. 
We vary the model size ($p$) when fitting the model from 3 to 1000, while keeping the underlying AR-3 process the same.
For each model size, ten time series were created with the same AR coefficients but with new random noise and both models are fitted to the same time-series.
Our experiments showed that \dar~ can fit to any sparse AR-process where the AR-coefficients are arbitrarily distributed over the $p$ lags. 
In the experiments, we have successfully fitted \dar~ with sparsity of up to $s=0.003$. 

\begin{figure}[htbp]
\centering
\includegraphics[height=0.35\linewidth]{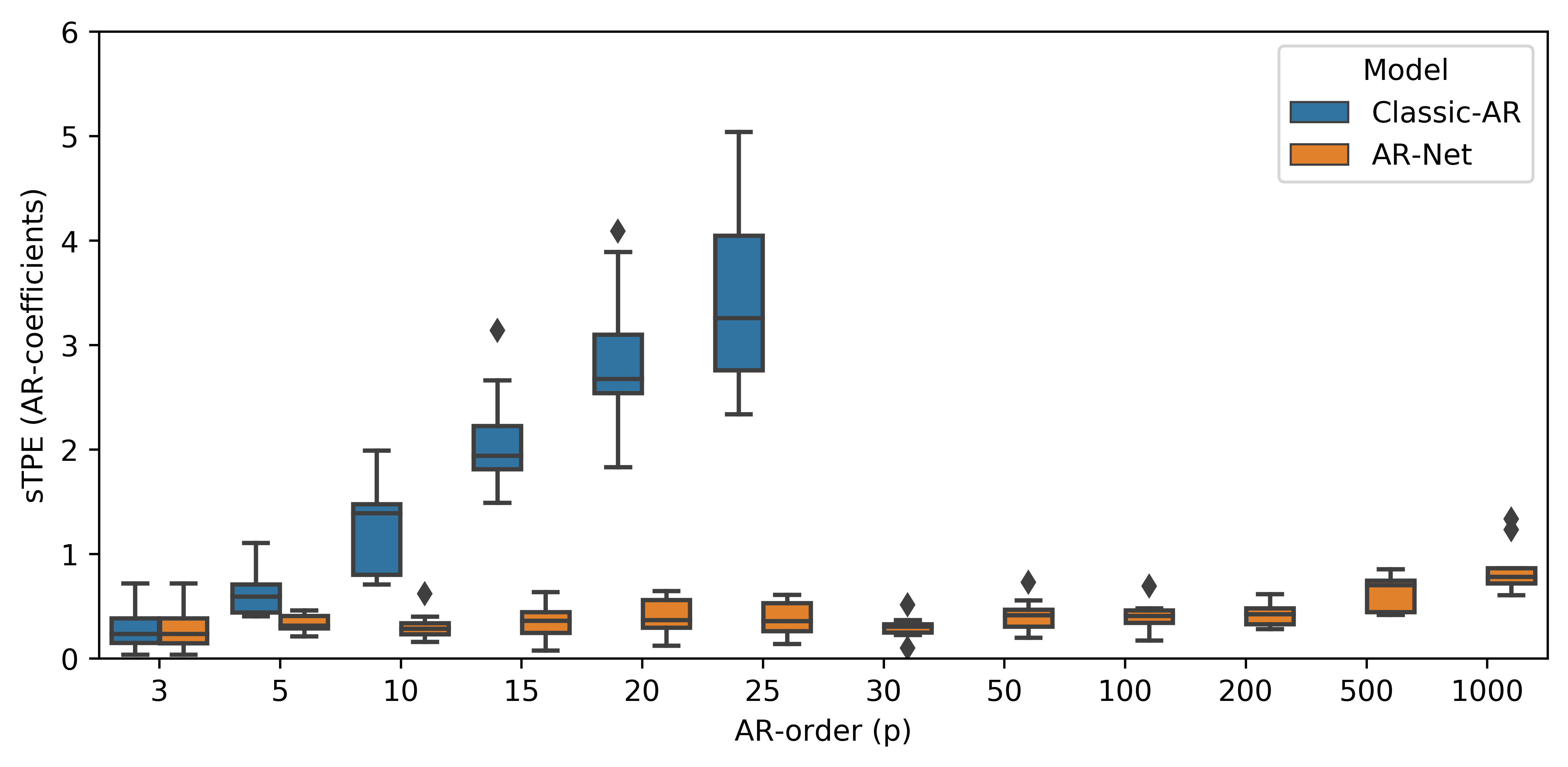}
\caption{Symmetrical Total Percentage Error of AR-coefficients weights for \car~ and \dar~ models for varying model sizes. Sparsity of the to be fitted AR-coefficients increases from 1.0 to 0.003 with the model size, as the data originates from an AR(3) process.}
\label{fig:sTPE_sparse}
\end{figure}

\begin{figure}[htbp]
\centering
 \includegraphics[height=0.35\linewidth]{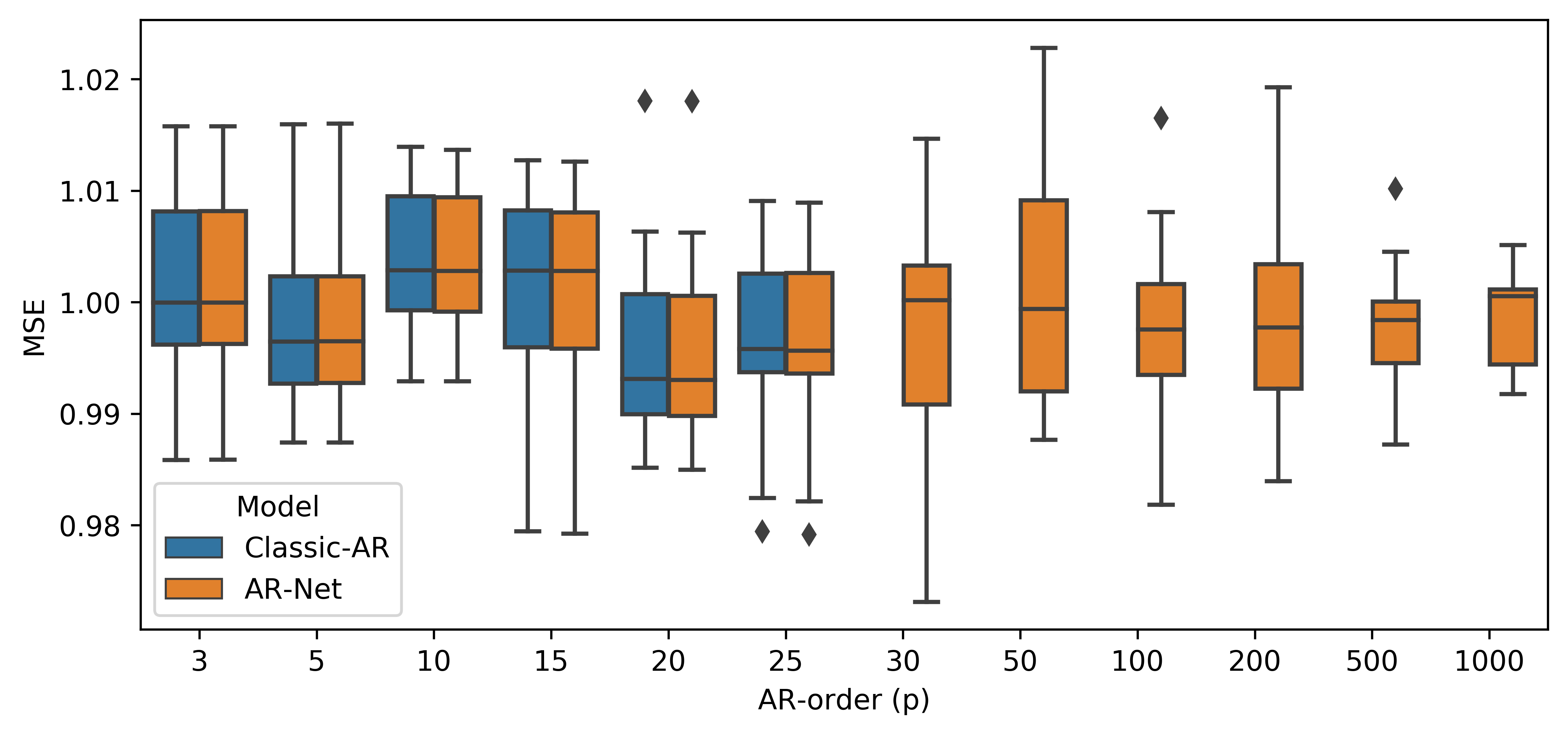}
\caption{Both \dar~ and \car~ effectively minimized the MSE loss term. Note that the dataset contains noise of one standard deviation.}
\label{fig:MSE_sparse}
\end{figure}

\begin{figure}[htb]
\centering
\includegraphics[width=0.8\linewidth]{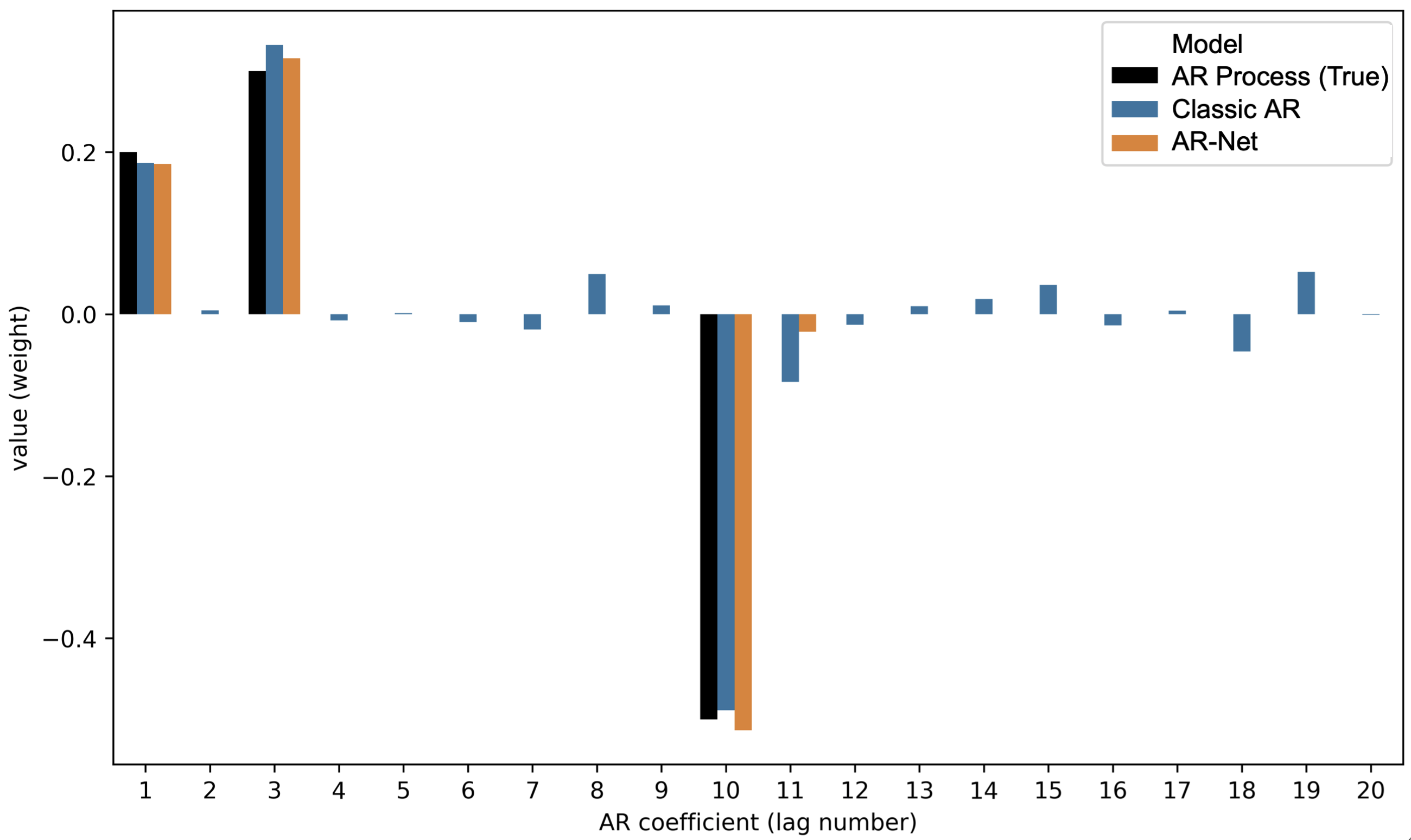}
\caption{Comparison of AR coefficients from  a \car~ and a sparse \dar~ model trained with $p=20$ on an AR-process of $p=3$ with sparsity. Only lags 1, 3 and 10 are non-zero. We used a small training dataset of 1000 samples with strong noise of one standard deviation.}
\label{fig:weights_sparse}
\end{figure}

From figure~\ref{fig:sTPE_sparse} we see that the precision of learned weights of \dar~ was superior to those of classic AR in all sparse scenarios. 
The sTPE of \car~ monotonically increased with $p$, while \dar~ remained precise up to a sparsity of $0.01$.  
While the \car~ model overfit to the noise in the dataset, the \dar~ model effectively learned the significant AR coefficients with little to no noise.
Nevertheless, both models achieved near identical forecasting performance (see figure~\ref{fig:MSE_sparse}).
Figure~\ref{fig:weights_sparse} shows an example of the weights learned by \dar~ and  \car~ on a sparse AR-process.

We acknowledge that this comparison was not entirely fair, as sparse implementation of \car~ exist. However, popular implementations of automatic sparsity, such as the addition of a spike and slab prior on the AR-coefficients, have an even greater computational complexity than the \car model evaluated here. Thus, they are not computationally tractable for AR processes with long range dependencies, which are the primary application of \dar.

\begin{figure}[htb]
\centering
\includegraphics[width=0.75\linewidth]{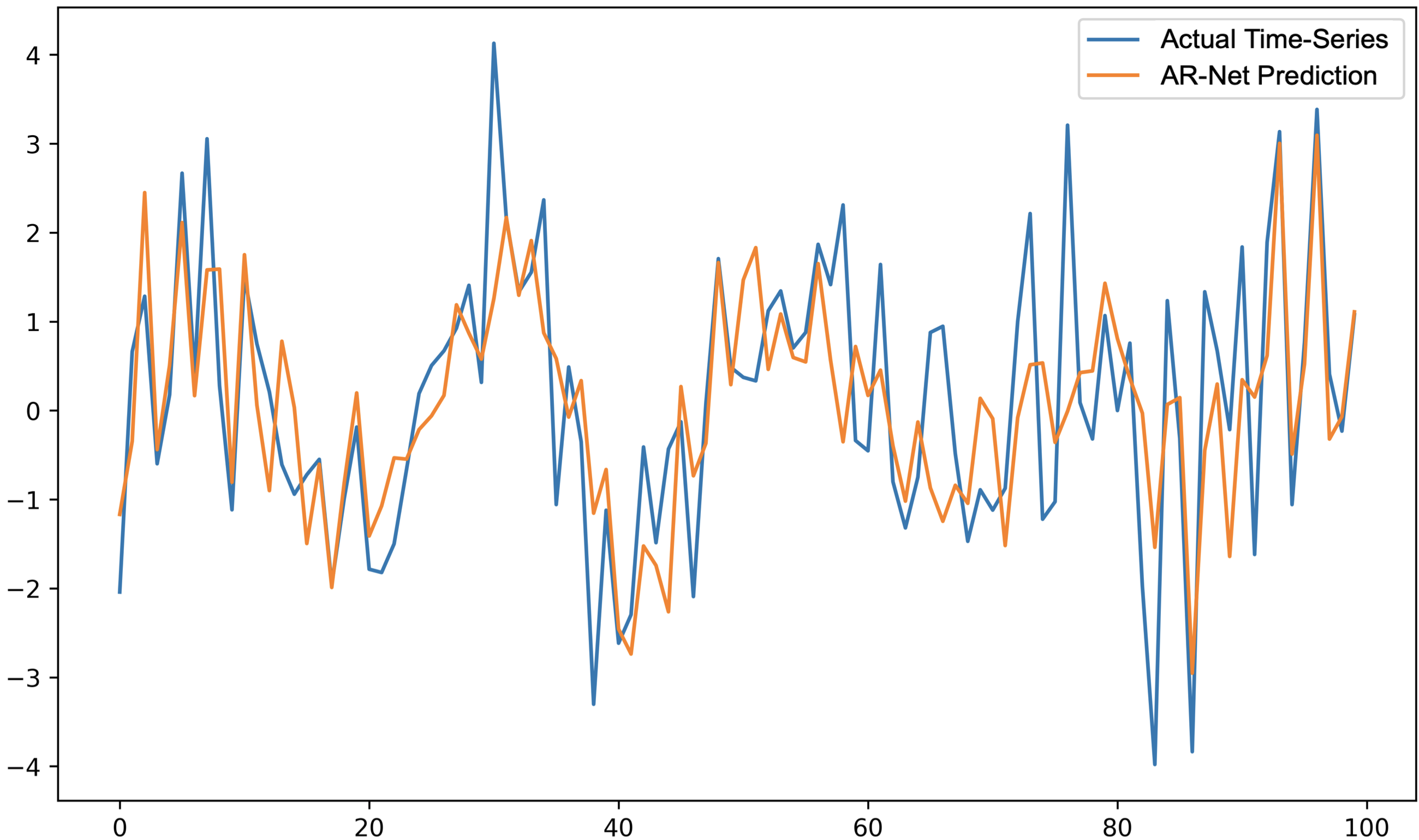}
  \caption{Prediction example from the test dataset for a sparse \dar~ model trained with $p=20$ on an underlying AR-process of $p=3$. The model was trained on a small times-series of only 1000 samples with strong noise (standard deviation of 1.0).}
  \label{fig:prediction_example}
\end{figure}

\paragraph{Small Datasets} are where statistics-based methods usually perform better than neural-network-based methods. 
We demonstrate that this is not the case for \dar with an experiment on a time-series with only 1000 samples with strong noise (std of 1.0). 
While both models achieved a similar MSE on the forecast, their sTPE of the learned AR-cofficients are vastly different, with 3.2 for \dar~  and 18.3 for \car. 
Figure~\ref{fig:weights_sparse} shows the results of fitting an AR(20) model on a sparse AR(3) process.
Figure~\ref{fig:prediction_example} shows an example of said model's predictions on the test dataset.

\newpage
\subsubsection{Computational Performance}
The run time complexity of least squares (used in \car) is $O(p^2 \cdot N)$ where $p$ is the number of features and $N$ is the training size. 
Here, $p$ is the order of the AR-process. 
For SGD (used in \dar), the complexity is roughly (only considering the forward pass) $O(E \cdot N / B \cdot \theta)$ where $E$ is the number of epochs, $B$ is the batch size and $\theta$ is the number of parameters.
$\theta$  is equivalent to number of input nodes times the number of output nodes, in our case $\theta = p \cdot 1$. 
Therefore, keeping batch size and number of epochs equivalent, we can see that as the AR-order $p$ increases, the SGD complexity remains linear with respect to $p$, while the fitting of least squares grows quadratically more complex as $p$ increases.
Thus, for high orders of $p$, \dar~ is clearly favorable compared to least squares. 

\begin{figure}[htbp]
\centering
\includegraphics[width=0.8\linewidth]{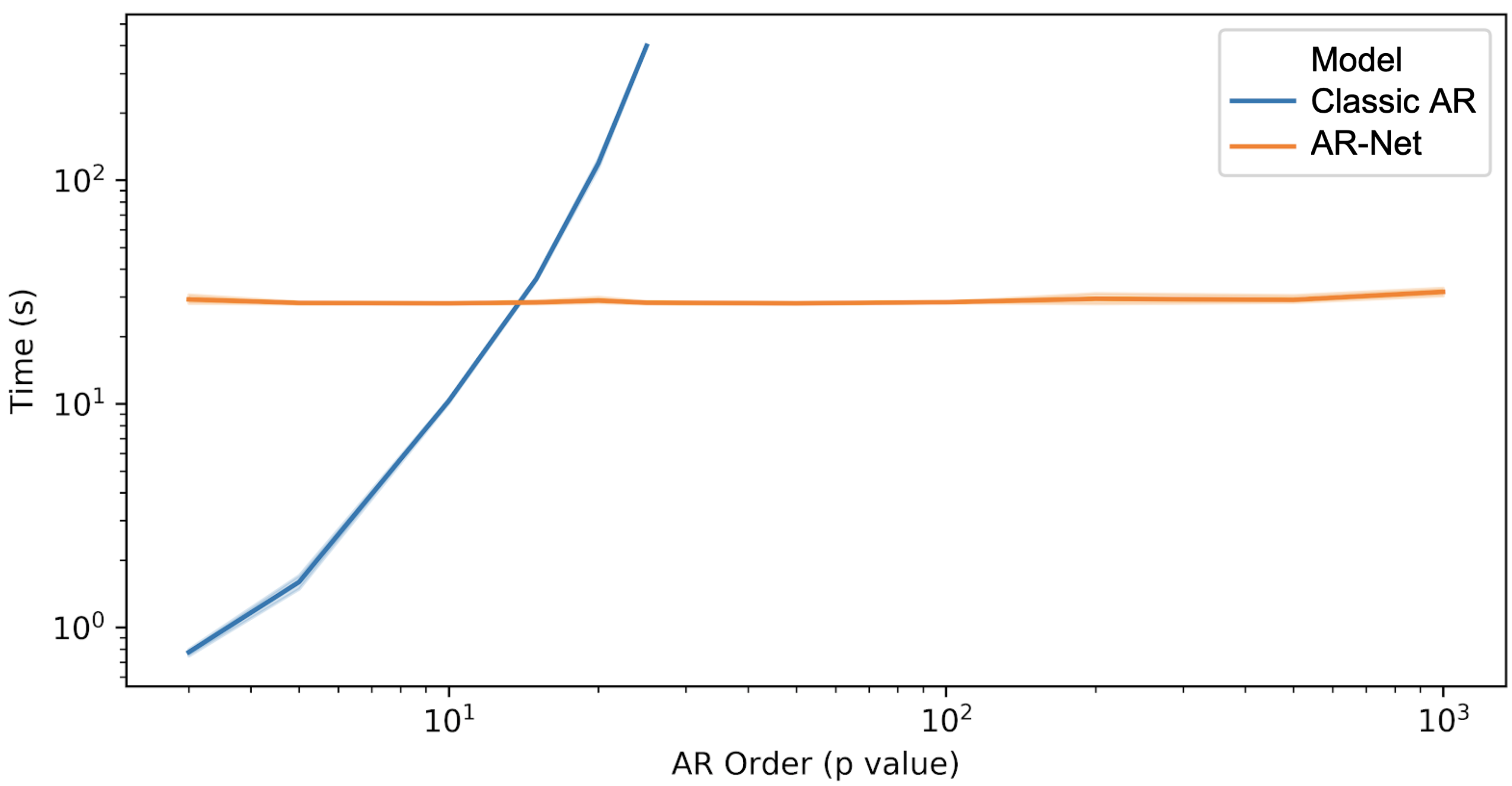}
\caption{Computation time increases linearly for \dar~ but quadratically for \car, as $p$ increases. Measured over five random runs for each order of p.}
\label{fig:comp-perf}
\end{figure}

We measure the actual training time in seconds for varying AR order $p$ in figure~\ref{fig:comp-perf}. 
\dar~ training was terminated after a fixed number of epochs, but it could have been sped up further by shortening the training time.
We found that for situations where both $N$ and $p$ are large, such as those prevalent in datacenter monitoring, \dar~ becomes the only computationally viable option, thanks to its linear computation time with respect to $p$.

\section{Conclusion}
\label{conclusion}

In this paper, we propose a \dar~ model, which uses stochastic gradient descent to estimate dynamics imposed by auto-regression. 
\dar~ makes it possible to learn a high order $p$ model orders of magnitude faster than using least squares. We show that the resulting weights are as interpretable as those of AR.  
Further, by adding regularization, \dar~ reliably selects and learns sparse weights, even up to a sparsity of $3:1000$. 
This eliminates the need to know the exact order of the AR-process and makes it possible to learn long-range dependencies on granular data without overfitting.

We found the sparse model to be insensitive to the estimated sparsity $s$ for estimates up to one magnitude off. 
However, as the model is trained with SGD, it is sensitive to learning rate and related hyperparameters. 
We hope to ease this sensitivity with a smart learning rate schedule such as the 1cycle-policy~\cite{1cycle}.

In future work, we will demonstrate how \dar~ makes it possible to seamlessly include co-variate time-series and to expand the forecast horizon, all with the same model. 
This makes it far simpler for the practitioner to expand their analysis from univariate one-step forecasting to multivariate multi-step forecasting.
Another part of our future work will be to extend \dar~ to have an \textit{MA} component and eventually include further temporal components (e.g., custom trend or seasonality).
Our long-term vision is to enable the practitioner with a simple but powerful time-series tool powered by neural networks.
 \paragraph{Acknowledgment} The work presented herein was funded in part by Total S.A in a research agreement with Stanford University. The views and opinions of authors expressed herein do not necessarily state or reflect those of the funding source.
\bibliographystyle{unsrt}  
\bibliography{other/references}

\begin{thebibliography}{10}

\bibitem{hyndman2014forecasting}
R.J. Hyndman and G.~Athanasopoulos.
\newblock {\em Forecasting: principles and practice}.
\newblock OTexts, 2014.

\bibitem{Ma_2017}
Yan Ma, Wenbing Zhu, Jinxia Yao, Chao Gu, Demeng Bai, and Kun Wang.
\newblock Time granularity transformation of time series data for failure
  prediction of overhead line.
\newblock {\em Journal of Physics: Conference Series}, 787:012031, jan 2017.

\bibitem{wireless_neural}
Gowrishankar Subrahmanyam and P~S.~Satyanarayana.
\newblock A time series modeling and prediction of wireless network traffic.
\newblock {\em International Journal of Interactive Mobile Technologies
  (iJIM)}, 3, 01 2009.

\bibitem{Hornik:1989:MFN:70405.70408}
K.~Hornik, M.~Stinchcombe, and H.~White.
\newblock Multilayer feedforward networks are universal approximators.
\newblock {\em Neural Netw.}, 2(5):359--366, July 1989.

\bibitem{Cybenko1989}
G.~Cybenko.
\newblock Approximation by superpositions of a sigmoidal function.
\newblock {\em Mathematics of Control, Signals and Systems}, 2(4):303--314, Dec
  1989.

\bibitem{prophetarticle}
Sean Taylor and Benjamin Letham.
\newblock Forecasting at scale.
\newblock {\em The American Statistician}, 72, 09 2017.

\bibitem{Sutskever:2014:SSL:2969033.2969173}
Ilya Sutskever, Oriol Vinyals, and Quoc~V. Le.
\newblock Sequence to sequence learning with neural networks.
\newblock In {\em Proceedings of the 27th International Conference on Neural
  Information Processing Systems - Volume 2}, NIPS'14, pages 3104--3112,
  Cambridge, MA, USA, 2014. MIT Press.

\bibitem{LSTM}
Sepp Hochreiter and J\"{u}rgen Schmidhuber.
\newblock Long short-term memory.
\newblock {\em Neural Comput.}, 9(8):1735--1780, November 1997.

\bibitem{attention}
Ashish Vaswani, Noam Shazeer, Niki Parmar, Jakob Uszkoreit, Llion Jones,
  Aidan~N. Gomez, Lukasz Kaiser, and Illia Polosukhin.
\newblock Attention is all you need, 2017.

\bibitem{wavenet}
Aaron van~den Oord, Sander Dieleman, Heiga Zen, Karen Simonyan, Oriol Vinyals,
  Alex Graves, Nal Kalchbrenner, Andrew Senior, and Koray Kavukcuoglu.
\newblock Wavenet: A generative model for raw audio, 2016.

\bibitem{NIPS2014_5346}
Ilya Sutskever, Oriol Vinyals, and Quoc~V Le.
\newblock Sequence to sequence learning with neural networks.
\newblock In Z.~Ghahramani, M.~Welling, C.~Cortes, N.~D. Lawrence, and K.~Q.
  Weinberger, editors, {\em Advances in Neural Information Processing Systems
  27}, pages 3104--3112. Curran Associates, Inc., 2014.

\bibitem{cond_wavenet}
Anastasia Borovykh, Sander Bohte, and Cornelis~W. Oosterlee.
\newblock Conditional time series forecasting with convolutional neural
  networks, 2017.

\bibitem{DBLP:journals/corr/CuiCC16}
Zhicheng Cui, Wenlin Chen, and Yixin Chen.
\newblock Multi-scale convolutional neural networks for time series
  classification.
\newblock {\em CoRR}, abs/1603.06995, 2016.

\bibitem{DBLP:journals/corr/abs-1903-02540}
Matteo Maggiolo and Gerasimos Spanakis.
\newblock Autoregressive convolutional recurrent neural network for univariate
  and multivariate time series prediction.
\newblock {\em CoRR}, abs/1903.02540, 2019.

\bibitem{LIME}
Marco~Tulio Ribeiro, Sameer Singh, and Carlos Guestrin.
\newblock "why should i trust you?": Explaining the predictions of any
  classifier, 2016.

\bibitem{SHAP}
Scott Lundberg and Su-In Lee.
\newblock A unified approach to interpreting model predictions, 2017.

\bibitem{chollet2015keras}
Fran\c{c}ois Chollet et~al.
\newblock Keras.
\newblock \url{https://keras.io}, 2015.

\bibitem{badexample}
Spyros Makridakis, Evangelos Spiliotis, and Vassilios Assimakopoulos.
\newblock Statistical and machine learning forecasting methods: Concerns and
  ways forward.
\newblock {\em PLOS ONE}, 13(3), 03 2018.

\bibitem{10.1371/journal.pone.0194889}
Spyros Makridakis, Evangelos Spiliotis, and Vassilios Assimakopoulos.
\newblock Statistical and machine learning forecasting methods: Concerns and
  ways forward.
\newblock {\em PLOS ONE}, 13(3):1--26, 03 2018.

\bibitem{10.1371/journal.pone.0211057}
Deepak~A. Kaji, John~R. Zech, Jun~S. Kim, Samuel~K. Cho, Neha~S. Dangayach,
  Anthony~B. Costa, and Eric~K. Oermann.
\newblock An attention based deep learning model of clinical events in the
  intensive care unit.
\newblock {\em PLOS ONE}, 14(2):1--17, 02 2019.

\bibitem{DBLP:journals/corr/ChePCSL16}
Zhengping Che, Sanjay Purushotham, Kyunghyun Cho, David~A. Sontag, and Yan Liu.
\newblock Recurrent neural networks for multivariate time series with missing
  values.
\newblock {\em CoRR}, abs/1606.01865, 2016.

\bibitem{Tang93feed-forwardneural}
Zaiyong Tang and Paul~A. Fishwick.
\newblock Feed-forward neural nets as models for time series forecasting.
\newblock {\em ORSA Journal of Computing}, 5:374--385, 1993.

\bibitem{1cycle}
Leslie~N. Smith.
\newblock A disciplined approach to neural network hyper-parameters: Part 1 --
  learning rate, batch size, momentum, and weight decay, 2018.

\bibitem{kour2014real}
George Kour and Raid Saabne.
\newblock Real-time segmentation of on-line handwritten arabic script.
\newblock In {\em Frontiers in Handwriting Recognition (ICFHR), 2014 14th
  International Conference on}, pages 417--422. IEEE, 2014.

\bibitem{ffvsts}
Zaiyong Tang and Paul Fishwick.
\newblock Feedforward neural nets as models for time series forecasting.
\newblock {\em INFORMS Journal on Computing}, 5:374--385, 11 1993.

\bibitem{DBLP:journals/corr/Gamboa17}
John Cristian~Borges Gamboa.
\newblock Deep learning for time-series analysis.
\newblock {\em CoRR}, abs/1701.01887, 2017.

\bibitem{boxjen76}
George.E.P. Box and Gwilym~M. Jenkins.
\newblock {\em Time Series Analysis: Forecasting and Control}.
\newblock Holden-Day, 1976.

\bibitem{DBLP:journals/corr/abs-1810-02281}
Sanjeev Arora, Nadav Cohen, Noah Golowich, and Wei Hu.
\newblock A convergence analysis of gradient descent for deep linear neural
  networks.
\newblock {\em CoRR}, abs/1810.02281, 2018.

\end{thebibliography}
\section{Appendix}
\label{appendix}
\subsection{Neural Networks Review}
Neural Networks (\textit{NN}) in their simplest form are composed of alternated layers of linear and non-linear functions, fitted to a target with stochastic gradient descent on a loss term \cite{hyndman2014forecasting}.
Stacking several layers in a "deep" neural network configuration allows to model complex nonlinear relationships between the response variable and its predictors \cite{hyndman2014forecasting}.
A neural network can be thought of as a network of `neurons' which are organised in layers. 
The predictors (or inputs) form the bottom layer, and the forecasts (or outputs) form the top layer. 
There may also be intermediate layers containing `hidden neurons' \cite{hyndman2014forecasting}.
\textit{NNs} have been widely used in economic, operational and financial fields \cite{kour2014real, ffvsts}.

\begin{figure}[ht]
\centering
\begin{subfigure}{.5\textwidth}
  \centering
  \includegraphics[width=.7\linewidth]{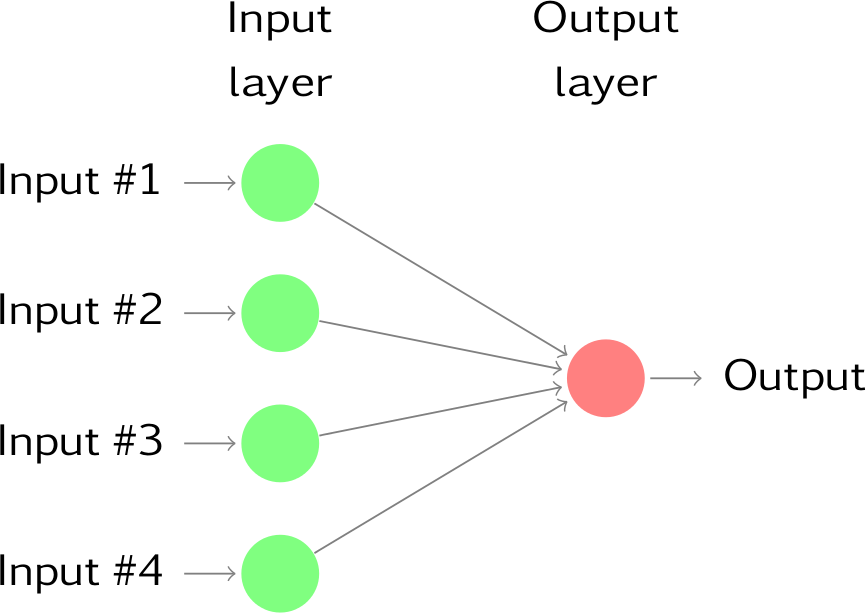}
  \caption{A simple neural network equivalent to a linear regression.}
  \label{fig:nn1}
\end{subfigure}%
\begin{subfigure}{.5\textwidth}
  \centering
  \includegraphics[width=.9\linewidth]{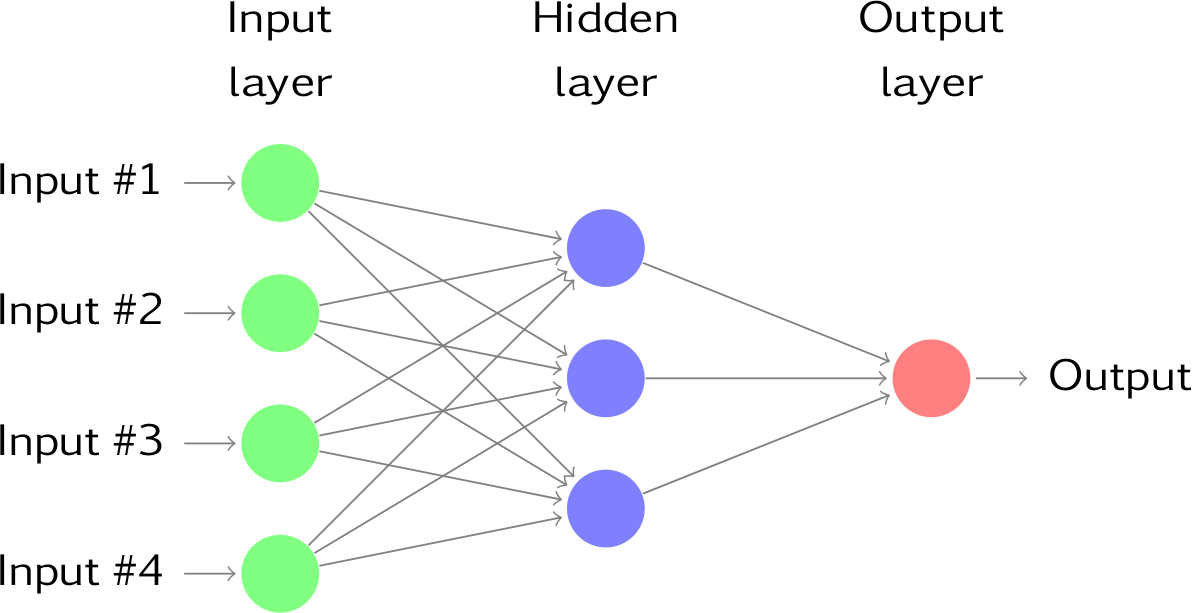}
  \caption{A neural network with  one hidden layer.}
  \label{fig:nn2}
\end{subfigure}
\caption{Visualization of neural networks.}
\label{fig:nn}
\end{figure}

The simplest neural networks contain no hidden layers, making it equal to a linear regression. Figure \ref{fig:nn1} \cite{hyndman2014forecasting} 
shows the neural network with four predictors. The coefficients attached to these 
predictors are called ``weights" and are used to obtain forecasts by a linear combination 
of the inputs.
The weights are learned in the neural network framework using gradient descent that minimizes a cost function
such as the MSE \cite{hyndman2014forecasting}.

Once we add an intermediate layer with hidden neurons, the neural network becomes non-linear. A simple example is shown in Figure \ref{fig:nn2}.
This is known as a multilayer feed-forward network, where each layer of nodes receives inputs from the previous layers. 
The inputs to each node are combined using a weighted linear combination. The result is then modified by a non-linear function, such as Sigmoid, $s(z)=\frac{1}{1+e^{-z}}$, or ReLU, $max(0,z))$, before becoming the input to the next layer. 

The weights of the neural network are ``learned" from the data. The parameter that restricts the weights is known as the ``regularizer", which is also often used to induce sparsity and to prevent the weights from becoming too large.

The weights are initialized with random values from some distribution, and are then updated  with SGD using the training data. Consequently, there is an element of randomness in the weights learned by a neural network, which is the result of each fitted model finding a \textit{different} suboptima.

\paragraph{Recurrent Neural Networks} 
RNN-based networks have loops that feed an output of the network as an input in the next time-stamp. 
By unfolding an RNN through time it is possible to train it using standard back-propagation. Unfolding a long network however may result in a vanishing gradient problem, causing the neural network to `forget' what happened a few steps behind, which is one of the reasons for developing LSTMs.
A hidden layer is replaced by a complex block of computing units composed of gates that manage forgetting and remembering 
of historical data \cite{DBLP:journals/corr/Gamboa17}. 

\subsection{ARMA Review} 
\label{ARMA-review}
The Auto-Regressive Moving-Average (\textit{ARMA}) model \cite{boxjen76} is one of the most widely used time-series forecasting methods in practice \cite{boxjen76}. An \textit{ARMA} model is derived from an \textit{AR} process, with an added moving average component. An \textit{ARMA}$(p,q)$ process, is parameterized by the orders $p$ and $q$ of the \textit{AR}$(p)$ and \textit{MA}$(q)$ components and their respective weights $w_i$ and $u_i$: 
$$y_t=c+\sum_{i=1}^{i=p}{w_i*y_{t-i}}+\sum_{i=1}^{i=q}{u_i*y_{t-i}}+e_t$$

\subsection{Fitting Procedures Review}

\paragraph{Error Back-propagation}
The error back propagation algorithm and its variations have been successfully used to train multilayer neural networks \cite{ffvsts}. The error back propagation consists of two processes through different layers of the neural network: a forward pass and a backward pass. In the forward pass the predicted model outputs are computed and then the errors, the difference between the measured outputs and the predicted outputs, are obtained. In the backward pass the error signals are used to update the weight/parameter estimates. Finally, each parameter is adjusted a step (of size learning rate) towards their respective impact on the minimized loss term. 
Gradient descent is guaranteed to converge to some optima, though not necessarily the global optima (see \cite{DBLP:journals/corr/abs-1810-02281} for more details, as the full proof is outside the scope of this paper).

\paragraph{Least Squares}
Least squares is a mathematical procedure for finding the best-fitting curve to a given set of points by minimizing the sum of the squares of the offsets (`the residuals') of the points from the curve. The sum of the squares of the offsets is used instead of the offset absolute values because this allows the residuals to be treated as a continuous differentiable quantity. However, because squares of the offsets are used, outlying points can have a disproportionate effect on the fit, a property which may or may not be desirable depending on the problem at hand. Estimating the least squares fit involves computation of a matrix inverse.

\end{document}